%% file: conference_101719.tex
\documentclass[conference]{IEEEtran}
\IEEEoverridecommandlockouts
\usepackage{cite}
\usepackage{amsmath,amssymb,amsfonts}
\usepackage{textcomp}
\usepackage{color,soul}
\usepackage{booktabs}
\usepackage{multirow}
\usepackage{subcaption}
\usepackage{xspace}
\usepackage{algorithm} 
\usepackage{algpseudocode}
\usepackage{balance}
\usepackage{lipsum}
\usepackage{graphicx}
\usepackage{caption}
\usepackage{hyperref}

\newcommand{\ie}{{\em i.e.,}\ }

\newcommand\mnamen{{\em Padding Module}\xspace}
\newcommand\mname{{\em \textit{Padding Module}}\xspace}
\newcommand\mnamet{{\em \textit{Padding Module}}}

\usepackage{amsmath}
\usepackage{graphicx}
\usepackage{algorithm}
\usepackage{algpseudocode}

\algdef{SE}[DOWHILE]{Do}{doWhile}{\algorithmicdo}[1]{\algorithmicwhile\ #1}%
\algnewcommand\algorithmicinput{\textbf{Input:}}
\algnewcommand\algorithmicoutput{\textbf{Output:}}
\algnewcommand\Input{\item[\algorithmicinput]}
\algnewcommand\Output{\item[\algorithmicoutput]}

\def\BibTeX{{\rm B\kern-.05em{\sc i\kern-.025em b}\kern-.08em
    T\kern-.1667em\lower.7ex\hbox{E}\kern-.125emX}}

\begin{document}

\title{\mnamen: Learning the Padding in Deep Neural Networks\\
\thanks{This paper has been accepted for publication by the IEEE Access.}
}

\author{\IEEEauthorblockN{Fahad Alrasheedi}
\IEEEauthorblockA{\textit{Department of Computer Science} \\
\textit{University of Nebraska Omaha}\\
Omaha, USA \\
falrasheedi@unomaha.edu}
\and
\IEEEauthorblockN{Xin Zhong}
\IEEEauthorblockA{\textit{Department of Computer Science} \\
\textit{University of Nebraska Omaha}\\
Omaha, USA \\
xzhong@unomaha.edu}
\and
\IEEEauthorblockN{Pei-Chi Huang}
\IEEEauthorblockA{\textit{Department of Computer Science} \\
\textit{University of Nebraska Omaha}\\
Omaha, USA \\
phuang@unomaha.edu}
% \and
% \and
% \and
}

\maketitle

\begin{abstract}
During the last decades, many studies have been dedicated to improving the performance of neural networks, for example, the network architectures, initialization, and activation. 
However, investigating the importance and effects of learnable padding methods in deep learning remains relatively open. 
To mitigate the gap, this paper proposes a novel trainable \mname that can be placed in a deep learning model. 
The \mname can optimize itself without requiring or influencing the model's entire loss function. 
To train itself, the \mname constructs a ground truth and a predictor from the inputs by leveraging
the underlying structure in the input data for supervision. 
As a result, the \mname can learn automatically to pad pixels to the border of its input images or feature maps. 
The padding contents are realistic extensions to its input data and simultaneously facilitate the deep learning model's downstream task.
Experiments have shown that the proposed \mname outperforms the state-of-the-art competitors and the baseline methods. For example, the \mname has 1.23\% and 0.44\% more classification accuracy than the zero padding when tested on the VGG16 and ResNet50. 
\end{abstract}


% \newpage
\input{tex/intro}
\input{tex/literature}

\input{tex/implemention}

\input{tex/results}

\input{tex/conclusion}

% here the refer
\bibliographystyle{IEEEtran}
\bibliography{refer}

\end{document}

%% file: tex/intro.tex
\section{Introduction}
\label{sec:intro}

Deep Neural Networks (DNNs) have significantly improved the performance of a wide range of computer vision tasks to the extent of being comparable to or exceeding human-level in many domains~\cite{RobustPhysical2}, such as image classification~\cite{he2019bag}, object recognition~\cite{speedAccuracy9}, and image segmentation~\cite{mask5}. 
DNNs for computer vision have been iteratively improving in different aspects such as network architecture~\cite{DeepConvolutional11, UniversalStyle15, ASimpleWay21,VeryDeep23}, network initialization~\cite{arpit2019initialize,Delving7}, optimization~\cite{Batch10,Adam12}, and activation~\cite{DeepResidual6,Densely8}. 
While it is intuitive that the salient foreground of an input image can control the results of a deep learning model~\cite{simonyan2013deep,zhou2016learning}, researchers have also discovered that the input's borders and corners can dominate the model's performance recently~\cite{innamorati2018learning,liu2018partial,distribution}. 
The study on the importance and effects of image borders remains relatively open, and this paper focuses on a trainable padding method that process image borders for deep learning models. 

Padding refers to the technique of adding extra data to the input's borders so that the input's width, height, or depth can be manipulated. 
Padding is widely used in Convolutional Neural Networks (CNNs) to alter the output size of a convolutional layer. 
Without padding, convolutional filters will not process the input’s borders and the output size will be reduced. 
The input size can be maintained with padding; we add an extra border before the convolution so that the original border can be processed~\cite{8308186}. 

\begin{figure}[t!]
\centering
\includegraphics[scale=0.25]{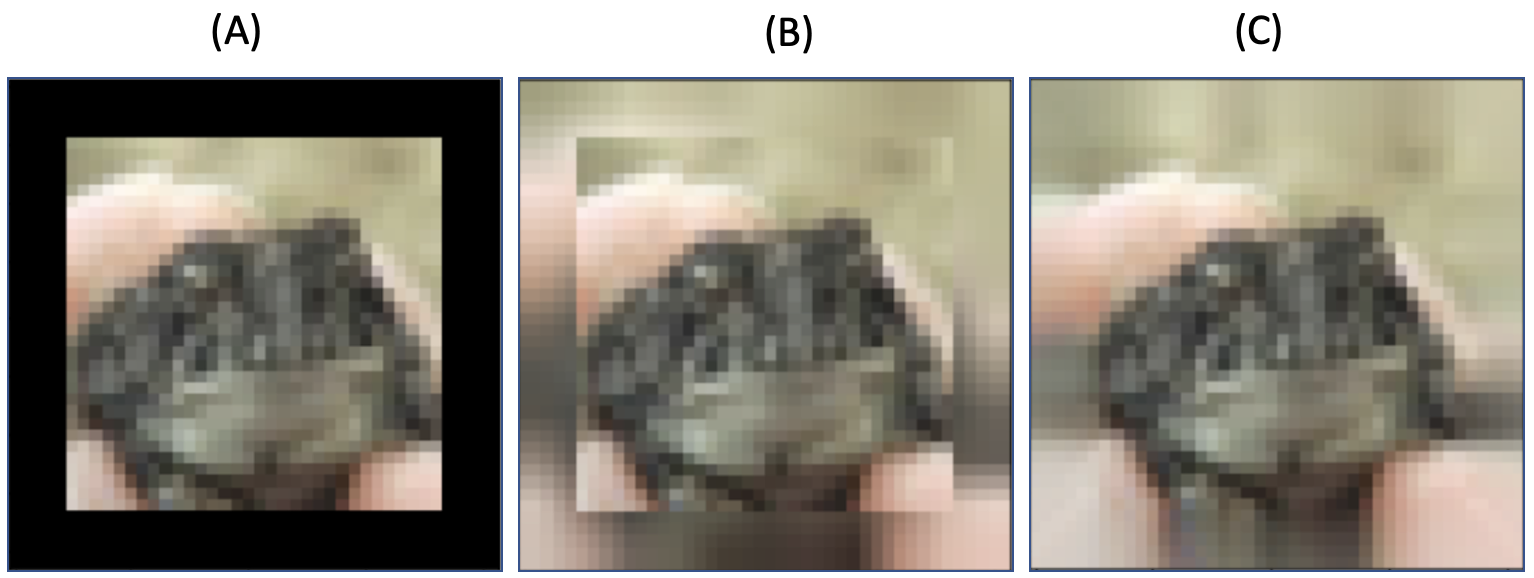} 
   \caption{Five-pixel padding applied to a CIFAR-10 sampled image using three different padding methods: A) the zero padding, B) the local mean interpolation, and C) the proposed \mname. 
   }
   \vspace{-2.0em}
   \label{fig:fig8}
\end{figure}

Traditional padding techniques include zero padding, replication padding, and reflection padding. 
The reflection padding reflects the valid data over the borders; 
the replication padding uses the borders themselves as padding values; 
the zero padding specifies the use of zeroes as padding values.
The replication and reflection padding methods extend the input with duplicate contents that may not be realistic; hence, they may destroy the original distribution~\cite{distribution}. 
The zero padding may outweigh the replication and reflection padding methods in terms of speed due to its computational simplicity. 
The major drawback of the traditional methods is that they are not dynamic.
Thus, the padding values are always static and not optimized during the model training in a way that how they could be optimally predicted rationally to the input's borders.

More recently, padding methods have been studied aiming at a more related and realistic extension of the original input~\cite{partial,distribution}. 
For example, Liu \textit{et al.}~\cite{partial} proposed a padding method using partial convolution. Nguyen \textit{et al.}~\cite{distribution} used a local mean of a sliding window over the input’s borders so the local distributions at the borders before and after the padding are consistent. 
These state-of-the-art padding methods outperformed the traditional padding in several tasks such as image classification, image segmentation, and image style transfer. 
However, the major disadvantage of the state-of-the-art padding methods is that they are not trainable: the padding contents are still not optimized.

In this paper, we propose a trainable \mname that can be inserted into selected positions of a deep learning model to learn how to pad its inputs.
The \mname can be trained simultaneously with a deep learning model, but, it is a self-learner in a way that will not require or influence the model’s entire loss function. 
During the training, the \mname internally constructs a ground truth from the input’s actual borders and trains a predictor considering the neighboring areas. 
The trained \mname can produce plausible results, as shown in Figure~\ref{fig:fig8}. 
The advantages of our work can be summarized as three-fold: 

\begin{itemize}
\item The proposed \mname introduces a trainable method that automatically pads its inputs. 
\item The \mname extends its input with realistic new data that are related to the original data. 
\item The \mname improves the performance of a downstream task of a deep learning model and outperforms the state-of-the-art competitors, \textit{e.g.}, classification. 
\end{itemize}

The remainder of this paper is organized as follows. In Section \ref{sec:literature}, we review the related work that addressed the padding effects on the neural networks performance and discuss how the current study fills the gap in the related work. Section \ref{sec:implem} discusses our approach for the \mname followed by evaluation results in Section \ref{sec:results}. Finally, Section \ref{sec:conclusion} concludes with the discussion on evaluation and highlights some of the future work in this sector.

%% file: tex/literature.tex
\section{Literature Review and Related Work}\label{sec:literature}
Many studies have tried to improve the performance of CNNs models from network architecture~\cite{he2016deep,huang2017densely,takikawa2019gated,radosavovic2020designing}, different variants of optimization~\cite{kingma2014adam,ioffe2015batch,liu2019variance}, activations~\cite{maas2013rectifier,he2015delving,ramachandran2017searching,mercioni2020p}, regularization methods~\cite{tompson2015efficient,larsson2016fractalnet}
and so no. However, little attention has been paid to investigating the padding schemes during the convolution operation. To assist the kernel, \ie features extractor, in extracting important features during image processing in CNNs, padding layers can be added to visit pixels of the images around the corners more times, and then increase accuracy. The previous padding methods are presented as follows: Section~\ref{subsec:per_imp_NN} presents the performance improvement of neural networks; Section~\ref{subsec:imp_space_design} introduces the improvement of space design; and Section~\ref{subsec:contributions} describes our contributions.

\subsection{Performance Improvement of Neural Networks}\label{subsec:per_imp_NN}
Several studies have proposed padding methods to improve the performance of the neural networks~\cite{partial,distribution,innamorati2018learning}.

Innamorati \textit{et al.}~\cite{innamorati2018learning} addressed the importance of the data at the borders of the input by proposing a convolution layer that dealt with corners and borders separately from the middle part of the image. They specifically designed filters for each corner and border to handle the data at the boundaries, including upper, lower, left, and right borders. The boundary filters used in the convolution were jointly learned with the filter used for the middle part of the image. However, the main issue of this study is that the number of filters used to deal with the boundaries increases linearly with the size of the receptive field.

Also, Nguyen \textit{et al.}~\cite{distribution} proposed a padding method that could keep the local spatial distribution of the padded area consistent with the original input. The proposed method used the local means of the input at the borders to produce the padding values; they proposed two different variants of the padding method: mean-interpolation and mean-reflection. Both variants used filters with static values, based on the receptive field, in the convolution operation that is supposed to yield the padding values maintaining the same distributions as the original borders. However, the main issue with this method is that they are not learnable.

Liu \textit{et al.}~\cite{partial} proposed a padding layer that uses a partial convolution that mainly re-weighted the convolution operation based on the ratio of the number of parameters in the filter to the number of valid data in the sliding window. In other words, they dealt with the padded area as hole areas that need to be in-painted, while the data coming from the original image were seen as non-hole areas. The main issue of this study is that the padding process is not learnable.

\subsection{Improvement of Spaces Design}\label{subsec:imp_space_design}
Also, some studies addressed the importance of the padding and data at the boundaries in the semantic representation learning and converting 360-degree space to 2-dimensional (2-D) space respectively~\cite{islam2020position, islam2021position, cheng2018cube}.
%This section provides an overview of the related work and how \mname fills the gap.  

Cheng \textit{et al.}~\cite{cheng2018cube} showed the importance of the padding method when they converted the 360-degree video to 2-dimensional space. They converted the video to six faces. Then, they used the reflection padding to connect them to form the 2-D space. The reflection padding naturally connected the faces compared to the zero-padding, which caused discontinuity.

Interesting works were provided by Islam \textit{et al.}~\cite{islam2020position, islam2021position} in which they showed the importance of zero padding along with the data at the borders in encoding the absolute position information in the semantic representation learning. They showed that the zero padding and the boundaries drove the CNN models to encode the spatial information that helped the filters where to look for a specific feature; the spatial information was eventually propagated over the whole image.

\subsection{Our Contributions}\label{subsec:contributions}
The padding methods and their effects on a CNN model’s performance are still open areas for researchers to investigate; hence, it is worth proposing new padding methods that could improve the performance of the CNN models. 
We propose a novel padding method, \mname, that could realistically extend the input with related data. It learns how to pad the input by using the inputs’ borders as a ground truth and the neighboring areas of the borders as a predictor. Then, it uses a local loss function such as Mean Squared Error (MSE) and updates the filters using the local differentiation of the loss function with respect to the \mname's filters. The following section explains the implementation of the \mname.

%% file: tex/implemention.tex
\section{The proposed \mnamen}
\label{sec:implem}
This paper presents the \mname, a learnable padding method that can pad the input with related and realistic padding, as shown in Figure~\ref{fig:fig8}. The \mname can be used as a substitute for other padding methods in the convolution layer, such as the zero padding, the replication padding, and the reflection padding. This section shows how the padding procedure (Section~\ref{sec:padding}) and the backpropagation (Section~\ref{sec:backprop}) of the \mname work.
 
\subsection{Padding Procedure}\label{sec:padding}
Algorithms \ref{forward} and \ref{backprop}, respectively, give an overview of the forward pass and the back-propagation of the \mname. The \mname first constructs a ground truth and a predictor from the input ( shown in step $1$ to step $3$ in Algorithm \ref{forward} and explained in Sections~\ref{sec:groundtruth} and~\ref{sec:predictors}). Then, the \mname uses the filters being learned to produce the actual padding values using the input’s borders as a predictor ( shown in steps $4$ to $13$ in Algorithm \ref{forward} and explained in Section~\ref{sec:actualPadding}). Finally, the \mname uses the MSE as a loss function to compute the loss value and updates the filters during the model's back-propagation ( shown in steps $1$ to $2$ in Algorithm \ref{backprop} and explained in Sections~\ref{sec:weights} and~\ref{sec:backprop}).

The \mname can pad the original input with any padding size, ($e.g.,$ one-pixel, two-pixels, etc). Indeed, the padding process in the \mname is iterative ( shown in steps $4$ to $13$ in Algorithm \ref{forward}). Assume the required padding size is three pixels, the padding process will iterate three times as follows: (1) padding the original input with one-pixel along all the four borders; (2) padding the output of the $1^{st}$ iteration with one-pixel along all the four borders; and (3) padding the output of the $2^{nd}$ iteration with one-pixel along all the four borders. Here, to easily explain our method, a simple case of padding process was presented here, $e.g.,$ one-pixel padding. Also, the \mname is assigned filters as many as the number of channels in the input as explained in Section~\ref{sec:weights}. Then, we explain the padding process considering a single channel. Here, the same procedure is separately applied to each channel in case of multiple channels.

\begin{figure*}[t]
   \centering
   \includegraphics[width = 0.9\linewidth]{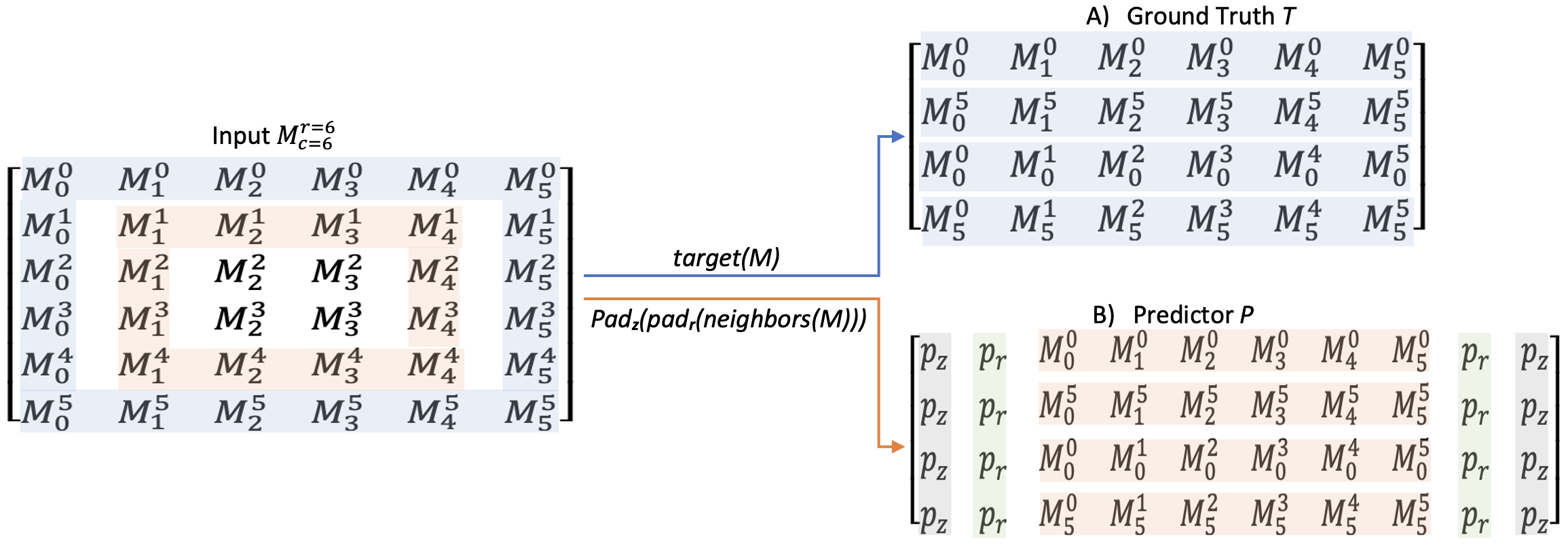}
   \caption{An example to illustrate the steps 1-3 in Algorithm~\ref{forward}. On the left: the input $M_{c}^{r}$ with size of $(6,6)$ pixels; the superscripts are the indexes in the row-wise traversal while the subscripts are the indexes in the column-wise traversal of the input. On the right: (A) the ground truth $T$: a result of applying step $1$ in Algorithm \ref{forward} which is a stack of the borders where the first row, second row, third row, and last row  are the upper, lower, left, and right borders in the input respectively; and (B) the predictor $P$: a result of applying steps $2$ and $3$ in Algorithm \ref{forward} which is a stack of the neighbors where the first row, second row, third row, and last row are neighbors to the upper, lower, left, and right borders in the input respectively, and the stack is padded at the left and right sides with reflection padding (\(p_{r}\)) and with zero padding (\(p_z\)). }
   \vspace{-1.0em}
   \label{fig:figureB}
\end{figure*}

\begin{algorithm}[htp]

\caption{Forward Pass}\label{forward}

\begin{algorithmic}[1]
  
  \Input{$M_{c}^{r}$, $size$, where $r$ and $c$ are the dimensions of a matrix, and $size$ is the padding size.}
  \Output{${M_{c'}^{r'}}$, where $r' = r + 2 \times size$, and  $c' = c + 2 \times size$.}
  \State $T\leftarrow target(M_{c}^{r})$ \hfill{/* as in Eq.\ref{mes_eqn2} */}
  \State $N\leftarrow neighbors(M_{c}^{r})$ \hfill{/* as in Eq.\ref{mes_eqn3} */}
  \State $P\leftarrow pad_{z}(pad_{r}(N))$ \hfill{/* as in Eq.\ref{mes_eqn4} */} 
  \State $M_{c'}^{r'}\leftarrow M_{c}^{r}$\hfill{/* initial state for $\overline{M_{c'}^{r'}}$  */}
  \While{$size\not=0$}
  \State $Nout\leftarrow borders(M_{c'}^{r'})$ \hfill{/* as in Eq.\ref{mes_eqn8} */}
  \State $Pout\leftarrow pad_{z}(pad_{r}(Nout)) $ \hfill{/* as in Eq.\ref{mes_eqn4} */}
  \State $O\leftarrow f_\theta(Pout)$ \hfill{/* as in Eq. \ref{mes_eqn9} */}
  \State $M_{c'+2}^{r'+2}\leftarrow O^{0} // pad_{z}(M_{c'}^{r'}) // O^{1}$ 
  \State $M_{c'+2}^{r'+2}\leftarrow sides( (O^{2})^{T} , M_{c'+2}^{r'+2}, (O^{3})^{T})$ \hfill{/* as in Eq.\ref{mes_eqn10} */}
  \State $M_{c'}^{r'}\leftarrow corners(M_{c'+2}^{r'+2})$ \hfill{/* as in Eq.\ref{mes_eqn11} */}
  \State $size\leftarrow size - 1$
  \EndWhile\label{euclidendwhile}
  \State \textbf{return} $M_{c'}^{r'}$
\end{algorithmic}
\end{algorithm}

\vspace{-1.5em}

\begin{algorithm}[htp]
\caption{Back Propagation}\label{backprop}
\begin{algorithmic}[1]  
  \Input{$G_{c'}^{r'}$, where $c'$ and $r'$ are the same dimensions of the output of the forward pass in Algorithm~\ref{forward}.}
  \Output{$G_{c}^{r}$, where $c$ and $r$ are the same dimensions of the input of the forward pass in Algorithm~\ref{forward}.}
  \State Compute the local gradients. \hfill{/* as in Eq.\ref{mes_eqn6} */}
  \State Update the filter weights 
  \State $G_{c}^{r}\leftarrow strip(G_{c'}^{r'})$ \hfill{/* as in Eq.\ref{mes_eqn12} */}
  \State \textbf{return} $G_{c}^{r}$
\end{algorithmic}
\end{algorithm}
% \vspace{-1.5em}

\vspace{+1.5em}
\subsubsection{Ground Truth $T$}\label{sec:groundtruth}
The \mname structures the ground truth $T$ by extracting the input's borders and stacking them upon each other vertically to form a four-row matrix. However, to stack the left and right borders vertically in $T$, they are transposed from column vectors to row vectors. Formally, given \(M_{c}^{r}\) as an original input with \(r\) and \(c\) as the number of rows and columns respectively; henceforth, superscripts and subscripts represent the indexes in the row-wise traversal and the column-wise traversal of the input respectively. The following is $T$'s extracting function \(target\) of the input \(M_{c}^{r}\):

\begin{equation} \label{mes_eqn2}
T= target(M_{c}^{r})= 
  \begin{bmatrix}
    M_{[:]}^{0}  \\
    M_{[:]}^{r-1}  \\
    {(M_{0}^{[:]})}^T  \\
    {(M_{c-1}^{[:]})}^T 
  \end{bmatrix}
  ,
 \end{equation}
where \(M_{[:]}^{0}\) is the entire row vector in \(M_{c}^{r}\) at index \(0\), \(M_{[:]}^{r-1}\) is the entire row vector in \(M_{c}^{r}\) at index \(r-1\), \({(M_{0}^{[:]})}^T\) is the transpose of the entire column vector in \(M_{c}^{r}\) at index \(0\), and \({(M_{c-1}^{[:]})}^T\) is the transpose of the entire column vector in \(M_{c}^{r}\) at index \(c-1\). Figure \ref{fig:figureB}. (A) is an example to visually illustrate how $T$ is constructed where the first row represents the upper border in \(M_{c}^{r}\), the second row represents the lower border in \(M_{c}^{r}\), the third row represents the left border in \(M_{c}^{r}\), and the last row represents the right border in \(M_{c}^{r}\).

\subsubsection{Predictor ($P$)}\label{sec:predictors}
To structure the predictor from the original input \(M_{c}^{r}\), the \mname extracts the row vectors that neighbor the upper border and lower border in \(M_{c}^{r}\) and the transpose of the column vectors that neighbor the left border and right border in \(M_{c}^{r}\). Then, the \mname stacks all the extracted neighbors vertically to form a four-row matrix. Formally, the predictor's (denoted as $P$) extracting function of \(M_{c}^{r}\) can be expressed in the following way:

First, the neighbors in \(M_{c}^{r}\) are selected and denoted as $N$ as follows:
\begin{equation} \label{mes_eqn3}
N= neighbors(M_{c}^{r})= 
  \begin{bmatrix}
    M_{[1:c-1]}^{1}  \\
    M_{[1:c-1]}^{r-2}  \\
    {(M_{1}^{[1:r-1]})}^T  \\
    {(M_{c-2}^{[1:r-1]})}^T 
  \end{bmatrix}
  .
 \end{equation}
The slice $[1:c-1]$ excludes the data in the row vectors at the borders due to overlapping with the $T$, whereas the slice $[1:r-1]$ excludes the data in the column vectors at the borders due to overlapping with $T$. 

Then, the \mname pads the structure as follows:
\begin{equation} \label{mes_eqn4}
P = pad_{z}(pad_{r}(N)).
 \end{equation}
First, the \(pad_{r}(.)\) function pads the structure with one pixel of the reflection padding horizontally (the left and right sides); then, with one pixel of the zero padding horizontally using the \(pad_{z}(.)\) function can get the final structure for $P$.

Each row in \(P\) will be used to predict the corresponding row in \(T\). For example, the first row in \(P\) will be used to predict the first row in $T$ representing the upper border in the input \(M_{c}^{r}\). Figure \ref{fig:figureB} (B) is an example to visually illustrate how the \mname constructs the stack of the neighbors (as a predictor) where the right and left sides of the stack are padded with the reflection padding (named as \textsl{\(p_{r}\)}), and the zero padding (named as \textsl{\(p_{z}\)}).

\subsubsection{Filters and the Loss Function}\label{sec:weights}
The \mname uses as many filters as the channels in the input (\ie filter per channel). Also, each filter will be a row vector with a size of $(1,3)$ and a stride of $(1, 1)$; that is because of having each row in \(P\) as a predictor for the corresponding row in $T$. Therefore, to predict $T$, the \mname convolutes the filters over \(P\); it uses its own loss function to optimize the prediction through the local differentiation of the loss function with respect to the filters. 

The loss function used by the \mname is the MSE which computes the squared difference between the ground truth and the predicted value. The following equation is the MSE's mathematical expression for a single data point: 
\begin{equation} \label{mes_eqn5}
MSE(f_{\theta}(P),T)= \sum_{a=1}^{4}\sum_{j=1}^{n}(\theta^{T} \cdot P^{a}_{j}- T^{a}_{j})^{2},
\end{equation}
where $f$ is the convolutional operation parameterized by ${\theta}$, $P$ and $T$ are the predictor and the ground truth extracted from the original input $M_{c}^{r}$, $a$ represents the indexes for rows in the four-row matrices $P$ and $T$, and $j$ represents the indexes for both the slide windows and columns in $P$ and $T$ respectively. Hence, $P^{a}_{j}$ is the $j$th slide window in the row indexed at $a$ in $P$, and $T^{a}_{j}$ is the corresponding value in $T$ indexed at the $a$th row and $j$th column.

The local differentiation of the \mname's loss function and the filters' updates are achieved during the model's back-propagation; these local gradients are not propagated to the previous layer. Besides that, the \mname  facilities the back-propagation of the model’s loss function going through it to the previous layer as  explained in Section~\ref{sec:backprop}. The following is the mathematical expression for the local gradients (\mname's loss function gradients with respect to a single filter for a single data point):

\begin{equation} \label{mes_eqn6}
\frac{\varphi}{\varphi\theta_{m}} MSE(f_{\theta}(P),T)= 2\sum_{a=1}^{4}\sum_{j=1}^{n}(\theta^{T} \cdot P^{a}_{j}- T^{a}_{j})x_{m},
\end{equation}
where \(x_{m}\) is a single feature in the \(P^{a}_{j}\) slide window which was multiplied by the corresponding weight, namely \(\theta_{m}\), in the $\theta$ during the convolution. %More importantly, during the model’s back-propagation, the padding module computes gradients of its loss function and updates its filters locally; these gradients are not propagated to the previous layer. However, the model facilities the back-pronation of the model’s loss function going through it to the previous layer.

%{Moreover, Section~\ref{sec:backprop} explains how the \mname back-propagates the model's gradients it receives to the previous layer. }

\subsubsection{Padding Process}\label{sec:actualPadding}

The procedures in Sections~\ref{sec:groundtruth},~\ref{sec:predictors}, and~\ref{sec:weights} are used to guide the \mname on learning how to predict the borders of the \textit{original} input, $M_{c}^{r}$, based on the neighboring areas to the borders, and then the \mname can optimize its filters. %as mentioned and explained in Section~\ref{sec:weights. 

However, the padding process is shown in steps $4$ to $13$ in Algorithm~\ref{forward}; it uses the borders of the input, $M_{c'}^{r'}$, as the predictor. In detail, the padding process iterates until the original input is padded with the required padding size. Hence, the original input $M_{c}^{r}$ is assigned to $M_{c'}^{r'}$ as an initial state in step $4$ before the padding loop starts. Then, each iteration pads the input, $M_{c'}^{r'}$, with one-pixel, and outputs a new $M_{c'}^{r'}$ which will be used for the next iteration and so forth. The dimensions of an iteration's output, $M_{c'}^{r'}$ in step $11$, are two-pixel larger than the dimensions of that iteration's input, $M_{c'}^{r'}$ in step $6$. 

Minutely, constructing the predictor in the padding process is similar to the way that constructs $P$ in Section~\ref{sec:predictors} with small modifications. To distinguish the notions of $neighbors$, $N$, and $P$, in Section~\ref{sec:predictors}, $borders$, $Nout$, and $Pout$ are denoted for the extracting function, the function's output, and the predictor, respectively. The following is the mathematical expression for the extracting function $borders$:

\begin{equation} \label{mes_eqn8}
Nout = borders(M_{c'}^{r'})= 
  \begin{bmatrix}
    M_{[:]}^{0}  \\
    M_{[:]}^{r-1}  \\
    {(M_{0}^{[:]})}^T  \\
    {(M_{c-1}^{[:]})}^T 
  \end{bmatrix}
  ,
 \end{equation}
where $M_{[:]}^{0}$ and $M_{[:]}^{r-1}$ mean extracting the entire upper and lower borders respectively. Whereas, $(M_{0}^{[:]})^T$ and $(M_{c-1}^{[:]})^T$ mean extracting the transpose of the entire left and right borders respectively.
Then, the \mname pads the output $Nout$ using Equation \ref{mes_eqn4} to get the final structure for $Pout$.

Consequently, convoluting the filters over the $Pout$ will produce the padding values for the iteration's input. The output can be expressed as follows:
\begin{equation} \label{mes_eqn9}
O= f_\theta(Pout),
\end{equation}
where $f$ is the convolutional operation parameterized by ${\theta}$, $Pout$ is the predictor, and the $O$ is the output and comes as a matrix of four rows. Each row represents the padding values for the corresponding area in the iteration's input, $M_{c'}^{r'}$, as follows: the first row ($O^0$), the second row ($O^1$), the third row ($O^2$), and the fourth row ($O^3$) represent the padding values for the upper, the lower, the left, and the right areas in the input respectively. 

Then, the steps from $9$ to $11$ are how the produced padding values stick around the input $M_{c'}^{r'}$. First, in step $9$, the vertical concatenation operator $//$ is used to concatenate the first row ($O^0$) with $M_{c'}^{r'}$, and then concatenates the resulted matrix with the second row ($O^1$). However, the rows from $O$ are two-pixel wider than the rows of $M_{c'}^{r'}$; therefore, to match the dimensions of these operands, the \mname uses $pad_{z}(.)$ to pad the $M_{c'}^{r'}$ horizontally with one pixel of the zero padding before the concatenation process. Hence, the  output's dimensions in step $9$, denoted as $M_{c'+2}^{r'+2}$, are two-pixel larger than the input $M_{c'}^{r'}$. Finally, the algorithm uses $sides$ function which can be formally expressed as the following:
\begin{equation} \label{mes_eqn10}
sides( (O^{2})^{T} , M_{c'+2}^{r'+2}, (O^{3})^{T}).
\end{equation}
This function does not change the dimensions; however, it adds respectively the transpose of the third row ($O^2$) and fourth row ($O^3$) to the left and right columns of $M_{c'+2}^{r'+2}$, the concatenated matrix with zero values at the left and right columns unless the corners already assigned values from the concatenation process. To resolve the double-count problem at the corners, the \mname takes the average of added values in the corners by dividing each corner by $2$; this averaging function is step $12$ in Algorithm~\ref{forward}:
\begin{equation} \label{mes_eqn11}
M_{c'}^{r'}= corners(M_{c'+2}^{r'+2}).
\end{equation}

Lastly, as mentioned early in this section that the dimensions of the iteration's output are two-pixel larger than the iteration's input. Hence, the output $M_{c'}^{r'}$, in Equation~\ref{mes_eqn11}, has dimensions $r'$ and $c'$ that are updated with the dimensions of $M_{c'+2}^{r'+2}$, namely $r'+2$ and $c'+2$ respectively.

\begin{figure*}[t]
   \centering
   \includegraphics[width=0.9\linewidth]{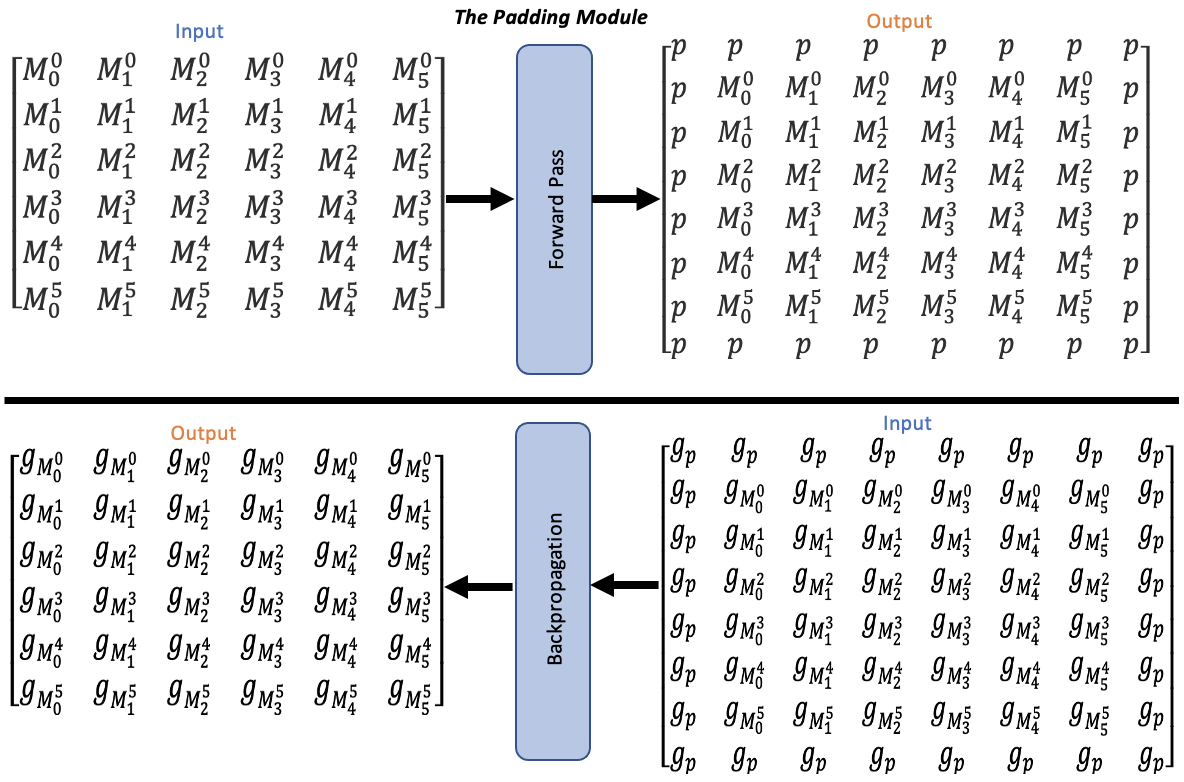}
   \caption{An example to illustrate the back propagation in Algorithm~\ref{backprop}. On the top: the input to the \mname is of size $(6,6)$, the \mname uses one-pixel padding and produces an output of size $(8,8)$ where the borders $p$ is the computed padding values. On the bottom: the back-propagation of the received gradients which is of size $(8,8)$ where the borders are gradients for the padding values $g_{p}$; the \mname strips out $g_{p}$ from the received gradients, and sends the remaining to the previous layer. The \textsl{$g$} stands for gradient; for example $ g_{M_{0}^{0}} $ is the gradient for the pixel at index $[0,0]$ in the input of the Forward Pass.}
   \vspace{-1.5em}
   \label{fig:figureC}
\end{figure*}

\subsection{Back-propagation}\label{sec:backprop}
As seen in Section~\ref{sec:weights}, the \mname is not optimized based on the model's main loss function; therefore, the model does not compute the gradients of its loss function with respect to the filters of the \mname. However, during the mode's backpropagation, the \mname achieves two key points as follows:
\begin{enumerate}
\item As shown in step $1$ in Algorithm \ref{backprop}, the \mname optimizes its filters through computing the local gradients for its loss function with respect to the filters as explained in Section~\ref{sec:weights}.
\item The process also receives $G_{c'}^{r'}$ which are the gradients of the model's loss function with respect to the \mname's output, the original input $M_{c}^{r}$ after being padded. Therefore, the \mname strips out the gradients from $G_{c'}^{r'}$ that represent the gradients for the padded areas in the \mname's output; the stripping-out process is step $3$ in Algorithm~\ref{backprop}, and formally expressed as follows:
\begin{equation} \label{mes_eqn12}
G_{c}^{r} = strip(G_{c'}^{r'}).
\end{equation}
Then, the \mname back-propagates to the previous layer the $G_{c}^{r}$, representing the gradients for the previous layer's output. Figure~\ref{fig:figureC} is an example to visually
illustrate how the back-propagation process in the \mname is achieved. 
\end{enumerate}

%% file: tex/results.tex
\section{Experimental Results and Analysis}\label{sec:results}
This section shows the design of the training and testing experiments on our \mname applied to a downstream task, \textit{i.e.}, image classification. The experimental setup is presented in Sections~\ref{subsec:exper_setup}. The quantitative and qualitative results are described in Section~\ref{subsec:quant_results} and~\ref{subsec:qual_results}.

\begin{figure*}
\centering
%    \fbox{%
\begin{minipage}[t]{0.49\textwidth}
  \centering
  \includegraphics[scale=0.27]{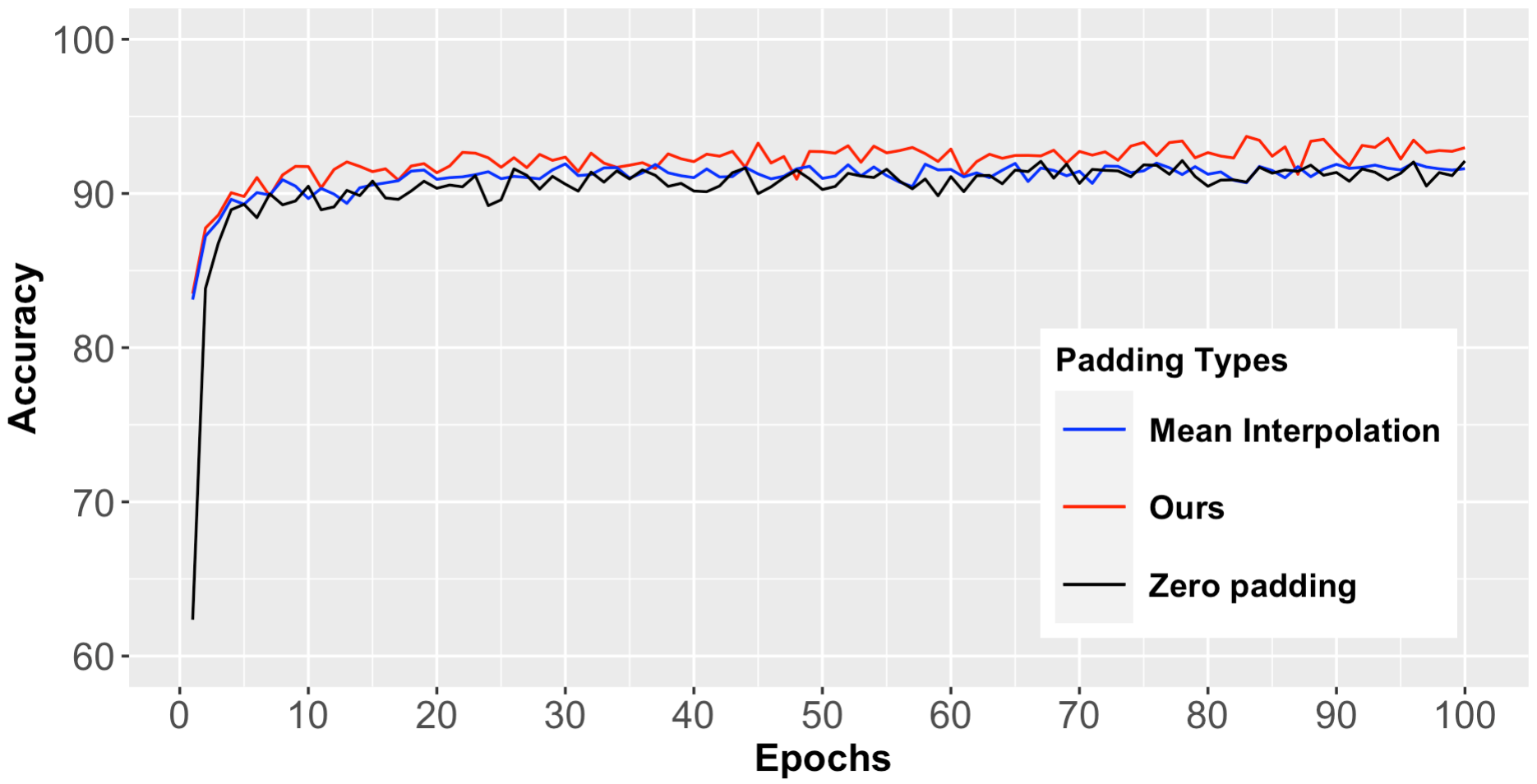}
  \captionof{figure}{The comparison of three different padding methods on the test images: zero padding, mean interpolation padding, and the \mname when applied to the VGG16 model.}
  \vspace{-1.5em}
  \label{fig:fig1}
\end{minipage}%
%    }
   \hfill
%    \fbox{%
\begin{minipage}[t]{0.49\textwidth}
  \centering
  \includegraphics[scale=0.27]{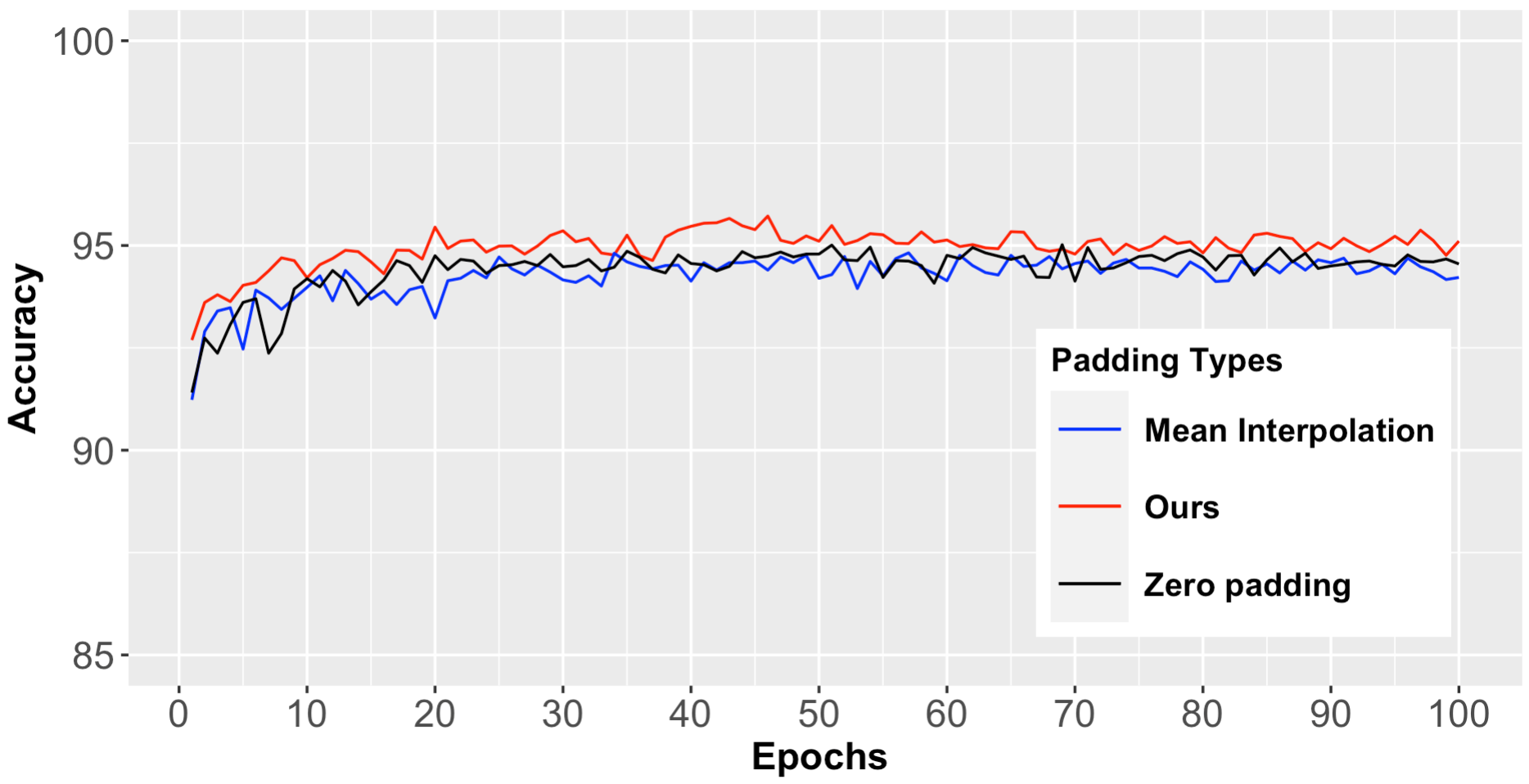}
  \captionof{figure}{The comparison of three different padding methods on the test images: zero padding, mean interpolation padding, and the \mname when applied to the ResNet50V2.}
  \vspace{-1.5em}
  \label{fig:fig2}
\end{minipage}
%    }
\end{figure*}

\subsection{Experiment Setup}\label{subsec:exper_setup}
The study used the premium service from Google Colaboratory where a GPU of Tesla T4 was assigned. The experiments and comparisons were conducted on the CIFAR-10 dataset for a classification task~\cite{cifar}. The CIFAR-10 dataset includes a training dataset of 50,000 images and a test dataset of 10,000 images. The images are of shape $(32, 32, 3)$, distributed equally to ten classes of airplane, automobile, bird, cat, deer, dog, frog, horse, ship, and truck. The \mname was applied to different networks namely: VGG16~\cite{VeryDeep23} and ResNet50V2~\cite{resnet50v2}; to make the deeper layers in these networks carry out a valid convolution, the images were resized to $(64, 64,3)$ and $(224, 224,3)$ for the VGG16 and the ResNet50V2, respectively. %\hl{We trained} \ph{Did you this sentence?}

The VGG16 is a vanilla-based architecture where the network shape is wider at the beginning of the network and narrowed down as going deep in the network. The pre-trained VGG16 was obtained from the keras\footnote{VGG16 from the keras:~\url{https://keras.io/api/applications/vgg}} without the top layers (the last three dense layers including the original softmax layer). Then, we added two fully-connected layers each with 512 neurons and followed by a dropout layer. On the other hand, the ResNet50V2 is made up of blocks where each block sends the block's input through the block itself, and also uses a skip connection to directly add the block's input to the output of the input's flow coming through the block. The process is known as the identity function that could help deep layers to improve the model's accuracy. 
ResNet50V2 is a modified version of the ResNet50~\cite{DeepResidual6}. The modification mainly is in the arrangement of the block layers; batch normalization~\cite{Batch10} and ReLU activation~\cite{relu} are applied to the data flow before the convolutional layer in the block. These changes enabled the ResNet50V2 to outperform the ResNet50 on the image classification task. The ResNet50V2 was downloaded from the keras\footnote{ResNet50V2 from the kera:~\url{https://keras.io/api/applications/resnet}} without the top layer (the last dense layer which is the original softmax layer). Then, two fully-connected layers with $1024$ and $512$ neurons were added.

Moreover, we added a softmax layer with ten outputs for both models of VGG16 and ResNet50V2, and then used the Adam optimizer \cite{Adam12} for the back-propagation of the gradients. Finally, the \mname was used before every convolutional layer in the VGG16; whereas, we replaced every zero padding layer in the ResNet50V2 with the \mname.

\subsection{Quantitative Results}\label{subsec:quant_results}
Section~\ref{subsubsec:image_classification_task} compares the proposed \mname and state-of-art padding solutions by performing the image classification task, and then Section~\ref{subsubsec:ablation_study} discusses an ablation study based on our solution.

\subsubsection{Image Classification Task}\label{subsubsec:image_classification_task}
We considered the zero padding method as a baseline to compare the \mname with.  Moreover, we used the mean interpolation padding method \cite{distribution} as the state-of-art since it outperformed the partial convolution padding method \cite{partial,distribution} in the image classification. The main goal of this study, which aligns with the literature, is to investigate the padding effect on the accuracy of DNN models. Therefore, the accuracy is used as a comparison metric between the performance of the \mname and the benchmark. The accuracy is the percentage of correctly classified images over the total number of images in the dataset.

Each model was trained with 100 epochs using the training dataset, and tested in each epoch using the test dataset. In Figure~\ref{fig:fig1}, the \mname outperforms both the baseline method and the mean interpolation padding method when using the VGG16; also, we found that the baseline is comparable to the mean interpolation method. As for the Resent50, the \mname also outperforms the other two paddings as shown in Figure \ref{fig:fig2}. We also noticed that the baseline method is comparable to the mean interpolation method. Moreover, Table \ref{table:20} summarizes the average of the last five epochs for the three different padding methods and the margin between the highest and the second-highest accuracies for the two models.

\begin{table}[h!]
  \begin{center}
    %\label{tab:table2}
    \begin {tabular*}{0.4\textwidth}{@{\extracolsep{\fill}} | c | c | c |  } %{tabular} {l|c|r} % <-- Alignments: 1st column left, 2nd middle and 3rd right, with vertical lines in between
      \hline 
      %\textbf{  & CNNs Models &}\\
      %\hline
      \textbf{Padding Method}& \textbf{VGG16 } & \textbf{ResNet50V2}  \\
      \hline
      \mname & 92.92 & 95.08 \\
      Zero Padding & 91.43 & 94.64 \\
      Mean Interoplation & 91.69 & 94.44 \\      
      \textbf{margin} & \textbf{1.23} & \textbf{0.44} \\
      \hline
    \end{tabular*}
  \caption{The average accuracy of the last five epochs for three different padding methods used in VGG16 and ResNet50V2. The margin is the difference between each model's highest and second-highest padding methods.}
  \vspace{-1.5em}
  \label{table:20}
  \end{center}
  \end{table}

\begin{figure}[t!]
  \centering
  \includegraphics[width=\linewidth]{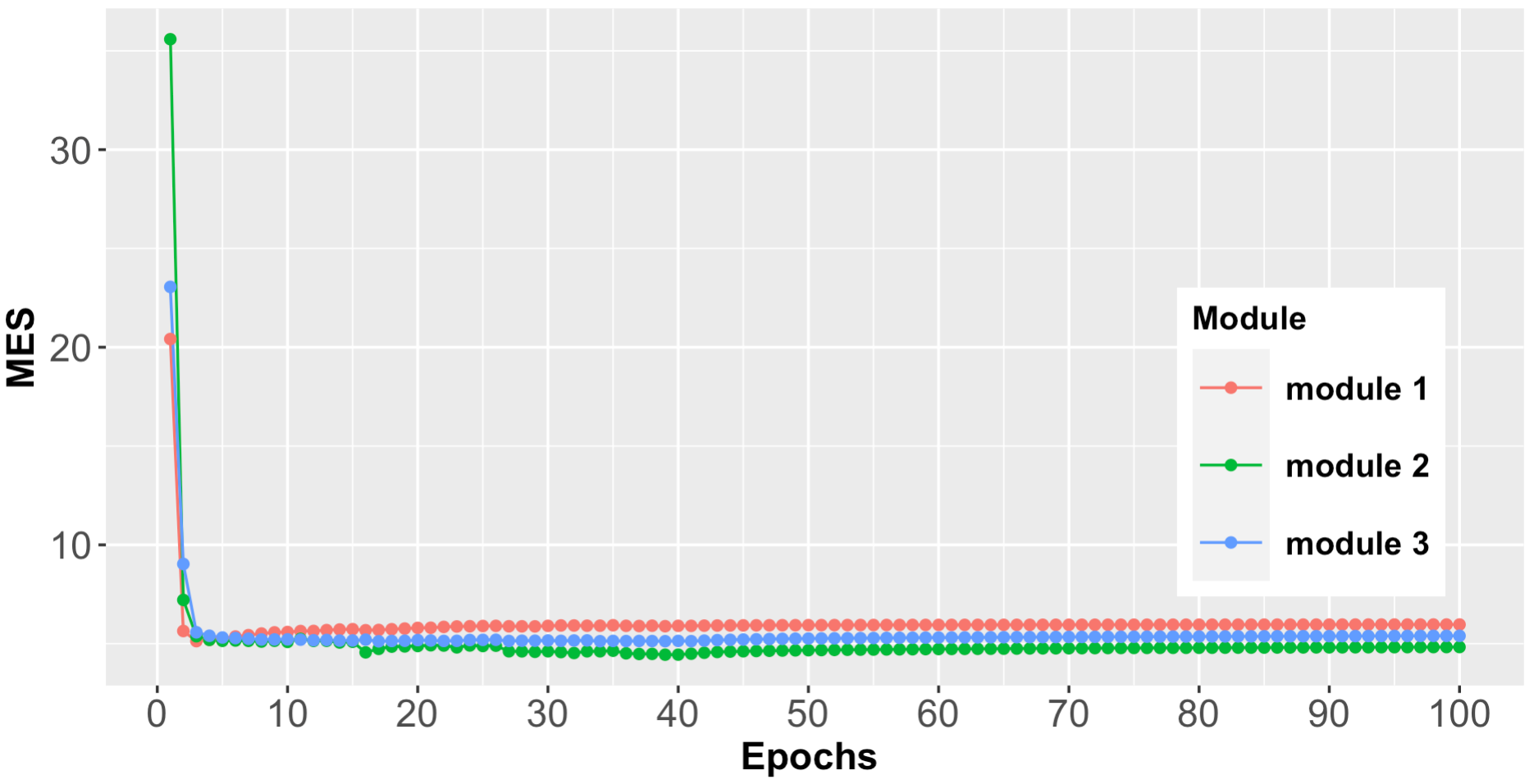}
  \caption{MSEs for three \mnamet s placed at different positions in the VGG16: 1) \textit{module 1}: at the beginning, 2) \textit{module 2}: at the middle, and 3) \textit{module 3}: at the end.}
  \vspace{-1.5em}
  \label{fig:mes}
\end{figure}

As mentioned, the study investigated the effects of the \mname on the accuracy of DNN models. Importantly, we showed that the related and realistic padding could improve the accuracy of DNN models ; therefore, the \mname was able to produce such padding by minimizing the MSE in Algorithms~\ref{forward} and~\ref{backprop}. Figure \ref{fig:mes} illustrates MSEs for three cases of the \mname applied in different places (at the beginning, in the middle, and at the end) in the VGG16; it is evident that the MSEs significantly decreased after only two epochs and then stayed flat till the end of the experiment for the three cases.

One natural drawback of the current \mname
was the extra running time caused by constructing
the data structures and optimizing the filters. Table 2 shows that VGG16 and ResNet50V2, on average, doubled the epoch's time when applying the \mname ($i.e.,$ placing the \mname before every convolutional layer in the VGG16 and replacing every zero padding layer in the ResNet50V2 with the \mname). On the other side, the accuracy for the VGG16 and ResNet50V2, respectively, gained margins of 1.49\% and 0.44\% when applying the \mname compared to the zero padding (no Padding Module). One remedy to lessen the running time problem may be to stop training the \mname when it significantly decreases the MSE after the first two epochs. However, improving the current \mname including the time complexity can be a further direction for further research.

\subsubsection{Ablation Study}\label{subsubsec:ablation_study}
\begin{figure*}%[!ht]
\centering
\includegraphics[scale=0.32]{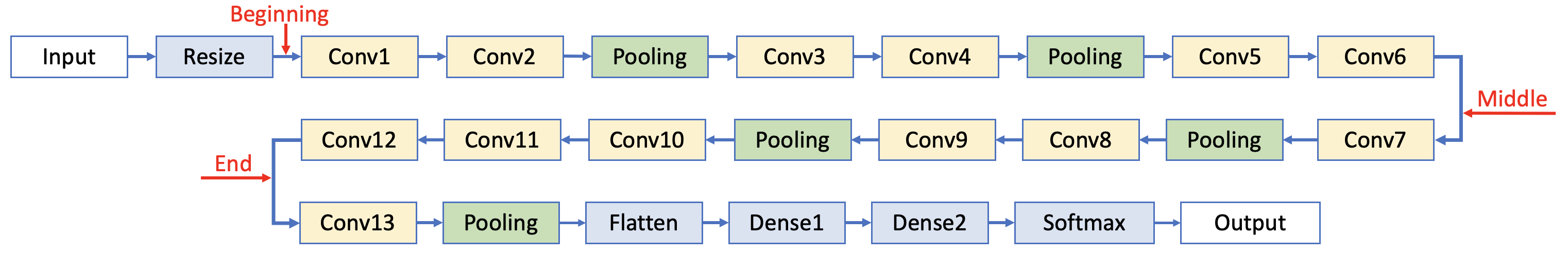} % ablation16
   \caption{The selected positions in the VGG16 for the \mname: at the beginning, middle, and the end.}
   \label{fig:fig3}
\end{figure*}

The experiments in this section were conducted as an ablation study where the \mname was empirically placed at different positions in the VGG16 model, as shown in Figure \ref{fig:fig3}: at the beginning of the model, in the middle, at the end, and the combination of all the three places together. We also compared the four scenarios with two other scenarios: (1) where the \mname was placed in all positions (before each convolutional layer) in the model; and (2) where the \mname was not used but the zero padding was used instead. We ran each scenario 100 epochs using the training dataset for training and the test dataset for evaluation, and averaged the test accuracies of the last five epochs for each scenario; Table~\ref{table:1} illustrates the summary of the comparison of the models. We noticed that using a single \mname with the shallow layers outperformed the case of using it with the deep layers. Also, the combination scenario showed a superiority
over the scenario of a single \mname. However, the best performance was when the \mname applied in the scenario of all positions. Finally, all the scenarios of applying the \mname outperformed the scenario of the model with no \mname.

\begin{table}[h!]
  \begin{center}
    %\label{tab:table2}
    \begin {tabular*}{0.461\textwidth}{@{\extracolsep{\fill}} | c | ll | ll | l|} %{tabular} {l|c|r} % <-- Alignments: 1st column left, 2nd middle and 3rd right, with vertical lines in between
    \hline
    \multirow{2}{*}{\textbf{Model}} & \multicolumn{2}{l|}{\textbf{Zero Padding}} & \multicolumn{2}{l|}{\textbf{Padding Module}} & \multirow{2}{*}{\textbf{Margin}} \\ \cline{2-5}
    & \multicolumn{1}{l|}{Accuracy} & Time & \multicolumn{1}{l|}{Accuracy} & Time & \\ \hline
    VGG16 & \multicolumn{1}{l|}{91.43}  & 2  & \multicolumn{1}{l|}{92.92}  & 4  & 1.49  \\ \hline
     ResNet50 & \multicolumn{1}{l|}{94.64}  & 5    & \multicolumn{1}{l|}{95.08}  & 9    & 0.44 \\ \hline
    \end{tabular*}
  \caption{On average, the running time doubles for one epoch when applying the \mname to the VGG16 and ResNet50V2 compared to the case of the zero padding (no \mname applied). Times are shown in a minute-scale. The margin is the accuracy difference between the case of applying the \mname and the zero padding. }
  \vspace{-1.0em}
  \label{table:2}
  \end{center}
  \end{table}

%%%%%%%%%%%%%%%%%%%%%%%%%%%

%%%%%%%%%%%%%%%%%%%%%%%%%%%

\begin{table}[h!]
  \begin{center}
    %\label{tab:table1}
    \begin {tabular*}{0.4\textwidth}{@{\extracolsep{\fill}} | c | l | r | } %{tabular} {l|c|r} % <-- Alignments: 1st column left, 2nd middle and 3rd right, with vertical lines in between
      \hline 
      \textbf{No}& \textbf{Different places in the VGG16} & \textbf{Accuracy} \\
      \hline
      1 & At the beginning & 91.98 \\
      2 & At the middle  & 91.8 \\
      3 & At the end & 91.97 \\
      4 & Combination of 1, 2, and 3 togather & 92.18 \\
      \textbf{5} & \textbf{All positions} & \textbf{92.8} \\
      6 & VGG16 with no \mname & 91.4 \\
      \hline
    \end{tabular*}
  \caption{Placing the \mname at different positions in the VGG16: 1) at the beginning 2) at the middle 3) at the end 4) combination of beginning, middle, end 5) before every convolutional layer 6) VGG16 with no \mname (zero padding instead).}
  \vspace{-2.0em}
  \label{table:1}
  \end{center}
  \end{table}

\begin{figure*}[t]
   \centering
   \includegraphics[width=520pt,height=250pt]{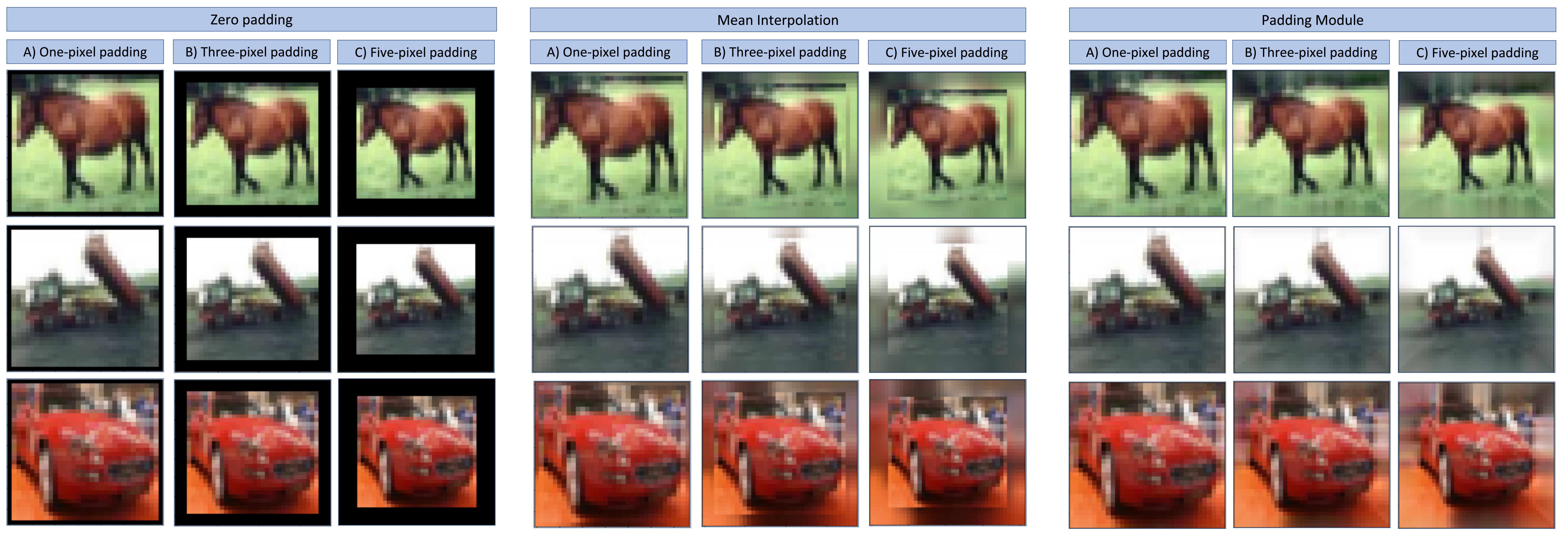}
   \caption{Three images sampled from CIFAR-10 and padded by different padding methods: zero-padding, mean interpolation, and the \mname. Each padding method uses three different padding sizes: A) one-pixel, B) three-pixel, C) five-pixel.}
   \label{fig:figurem}
\end{figure*}

\subsection{Qualitative Results}\label{subsec:qual_results}
Different padding sizes, such as one-pixel, three-pixel, and five-pixel, were used to illustrate how the \mname can extend the input with related and realistic extensions. Also, we compared these different padding sizes with the other two methods, namely the zero padding and the mean interpolation padding. As shown in Figure~\ref{fig:figurem}, the \mname can learn how to pad the input with related data and natural extension; this finding becomes more evident as the padding size increases.

%% file: tex/conclusion.tex
\section{Future Research Directions and Conclusion}\label{sec:conclusion}
This paper proposed a novel padding method: \mname; that can learn how to pad an input from the input's borders; hence, the input can be realistically extended with related data. The \mname is a self-learning of its weights. To train itself, the \mname constructs a ground truth and a predictor from the inputs by leveraging the underlying structure in the input data for supervision. The \mname uses convolutional operation over the predictor to produce a predicted value that is, in turn, compared with the ground truth. The \mname uses a local loss function, independent from the model's main loss function, to minimize the difference between the predicted value and the ground truth. Therefore, the \mname updates its convolutional filters locally during the model's back-propagation. Besides that, the \mname back-propagates the model's gradients with respect to the \mname's output after stripping out the gradients for the padded areas to the previous layer.

\iffalse
The \mname is assigned a separate loss function from the model's main loss function, and the \mname's loss function is minimized during the model's backpropgation independently of the model's main loss function. To optimize the \mname's loss function, the \mname constructs a ground truth and a predictor from the input, and then it applies convolutional filters over the predictor to produce a predicted value that is, in turn, compared with the ground truth. Hence, the difference between the ground truth and the predicted value is minimized using the \mname's local gradients during the model's back-propagation. Besides that, the \mname receives \hl{the model's gradients}\ph{are you saying "gradients with respect to a model's trainable variables"?} and back-propagates them to the \hl{previous layer} \ph{more precise which layer}.
\fi

The experimental results showed that the \mname outperformed the zero-padding and the state-of-art padding in the image classification task. In the ablation study, we also observed that using a single \mname with the shallow layers improved the performance slightly better than using it with the deep layers in the VGG16 network. On the other hand, using three of the \mname placed in different positions (at the beginning, at the middle, and at the end) in the VGG16 outperformed the scenario of a single \mname. Moreover, placing the \mname in all positions (before every convolutional layer) in the VGG16 outperformed all other scenarios as shown in Table~\ref{table:1}.

Our experiments applied the \mname to the two well-known networks: VGG16 and ResNet50, for the image classification task. The VGG16 and ResNet50 networks were chosen to represent small and large networks, respectively. They, also, were used by the literature; hence, we used them to compare the \mname with the previous work. Although two different networks are only used in one task, we shall extend the \mname to improve such networks in different tasks, including object detection, style transfer, and image inpainting. We leave investigating the \mname in a wide range of tasks for future research.

Also, the \mname learned how to pad the input independently of the model's loss function. However, it is possible to optimize the \mname's filters based on optimizing the model's main loss function; this approach will be entirely different. Hence, one research direction may be to implement a padding method that can optimize its padding filters based on the model's main loss function.